\pdfoutput=1
\documentclass[11pt]{article}
\usepackage[namelimits]{amsmath}
\usepackage{ACL2023}

\usepackage{times}
\usepackage{latexsym}
\usepackage{geometry}
\usepackage{colortbl}
\usepackage{color}
\usepackage{algorithmicx,algorithm}
\usepackage{arydshln}
\makeatletter
\renewcommand{\maketag@@@}[1]{\hbox{\m@th\normalsize\normalfont#1}}%

\makeatother

\usepackage[T1]{fontenc}
\usepackage[utf8]{inputenc}

\usepackage{microtype}
\usepackage{graphicx}
\usepackage{inconsolata}
\usepackage{booktabs}
\usepackage{multirow}
\usepackage{enumitem}
\usepackage{amsmath}
\usepackage{amssymb}
\usepackage{booktabs}
\usepackage{float}
\usepackage{xurl}
\usepackage{subcaption}
\usepackage{amsmath}
\usepackage{multirow}
\usepackage{makecell}
\usepackage{booktabs}
\usepackage{bm}

\usepackage{bbm}
\usepackage{amssymb}
\usepackage{adjustbox}
\usepackage{graphicx} %插入图片的宏包
\usepackage{float} %设置图片浮动位置的宏包
\usepackage[noend]{algpseudocode}
\usepackage{algorithmicx,algorithm}
\usepackage{microtype}
\usepackage{array}
\DeclareMathAlphabet\mathbfcal{OMS}{cmsy}{b}{n}
\title{ML-LMCL:~Mutual Learning and Large-Margin Contrastive Learning for Improving ASR Robustness in Spoken Language Understanding}

\author{ Xuxin Cheng, Bowen Cao\textsuperscript{\rm \textdagger}, Qichen Ye\textsuperscript{\rm \textdagger}, \\ \textbf{Zhihong Zhu\textsuperscript{\rm \textdagger}, Hongxiang Li, Yuexian Zou\textsuperscript{{\rm *}}}
	\\School of ECE, Peking University, China\\
	\{chengxx, cbw2021, zhihongzhu, lihongxiang\}@stu.pku.edu.cn\\
	\{yeeeqichen, zouyx\}@pku.edu.cn}
 
\usepackage{lipsum}
\newcommand\blfootnote[1]{%
  \begingroup
  \renewcommand\thefootnote{}\footnote{#1}%
  \addtocounter{footnote}{-1}%
  \endgroup
}

\begin{document}
\maketitle
\blfootnote{\textsuperscript{\rm \textdagger} Equal contribution.}
\blfootnote{\textsuperscript{{\rm *}} Corresponding author.}
\begin{abstract}
Spoken language understanding~(SLU) is a fundamental task in the task-oriented dialogue systems. However, the inevitable errors from automatic speech recognition~(ASR) usually impair the understanding performance and lead to error propagation. Although there are some attempts to address this problem through contrastive learning, they~(1)~treat clean manual transcripts and ASR transcripts equally without discrimination in fine-tuning;~(2)~neglect the fact that the semantically similar pairs are still pushed away when applying contrastive learning;~(3)~suffer from the problem of Kullback–Leibler~(KL) vanishing. In this paper, we propose \textbf{M}utual \textbf{L}earning and \textbf{L}arge-\textbf{M}argin \textbf{C}ontrastive \textbf{L}earning~(ML-LMCL), a novel framework for improving ASR robustness in SLU. Specifically, in fine-tuning, we apply mutual learning and train two SLU models on the manual transcripts and the ASR transcripts, respectively, aiming to iteratively share knowledge between these two models. We also introduce a distance polarization regularizer to avoid pushing away the intra-cluster pairs as much as possible. Moreover, we use a cyclical annealing schedule to mitigate KL vanishing issue. Experiments on three datasets show that ML-LMCL outperforms existing models and achieves new state-of-the-art performance.
\end{abstract}

\section{Introduction}
\begin{figure}[t]
\begin{minipage}[a]{1.0\linewidth}
  \centering
\centerline{\includegraphics[width=7.75cm]{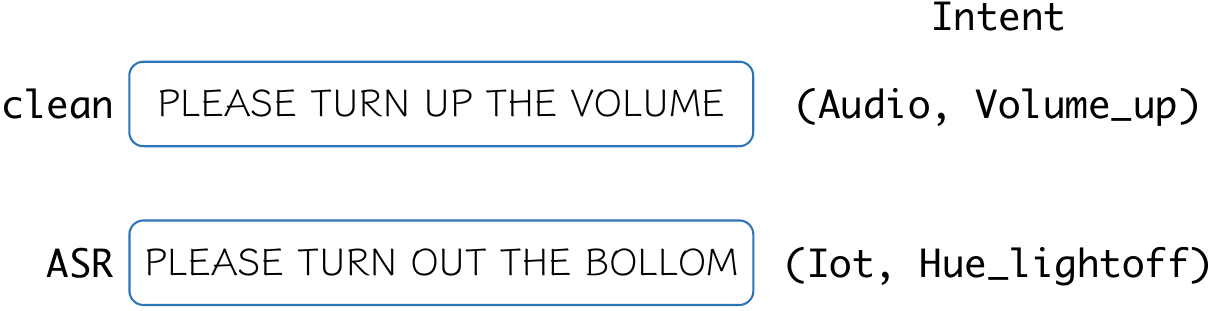}}
  \smallskip
\end{minipage}
\caption{An example of the intent being predicted incorrectly due to the ASR error.}
\label{fig:example}
\end{figure}

Spoken language understanding(SLU) is an important component of various personal assistants, such as Amazon's Alexa, Apple's Siri, Microsoft's Cortana and Google's Assistant~\citep{young2013pomdp}. SLU aims at taking human speech input and extracting semantic information for two typical subtasks, mainly including intent detection and slot filling~\citep{tur2011spoken}. Pipeline approaches and end-to-end approaches are two kinds of solutions of SLU. Pipeline SLU methods usually combine automatic speech recognitgion~(ASR) and natural language understanding~(NLU) in a cascaded manner, so they can easily apply external datasets and external pre-trained language models. However, error propagation is a common problem of pipeline approaches, where an inaccurate ASR output can theoretically lead to a series of errors in subtasks. As shown in Figure \ref{fig:example}, due to the error from ASR, the model can not predict the intent correctly. Following \citet{chang22c_interspeech}, this paper only focuses on intent detection. 

Learning error-robust representations is an effective method to mitigate the negative impact of errors from ASR and is gaining  increasing attention. The remedies for ASR errors can be broadly categorized into two types: (1) applying machine translation to translate the erroneous ASR transcripts to clean manual transcripts~\citep{mani2020asr,wang2020asr,dutta2022error}; (2) using masked language modeling to adapt the model. However, these methods usually requires additional speech-related inputs~\citep{huang2019adapting,sergio2020attentively,wang2022arobert}, which may not always be readily available. Therefore, this paper focuses on improving ASR robustness in SLU without using any speech-related input features.

Despite existing error-robust SLU models have achieved promising progress, we discover that they suffer from three main issues:

(1)~\textbf{Manual and ASR transcripts are treated as the same type.} In fine-tuning, existing methods simply combine manual and ASR transcripts as the final dataset, which limits the performance. Intuitively, the information from manual transcripts and the information from ASR transcripts play different roles, so the model fine-tuned on their combination cannot discriminate their specific contributions. Based on our observations, models trained on the clean manual transcripts usually has higher accuracy, while models trained on the ASR transcripts are usually more robust to ASR errors. Therefore, manual and ASR transcripts should be treated differently to improve the performance of the model.

(2)~\textbf{Semantically similar pairs are still pushed away.} Conventional contrastive learning enlarges distances between all pairs of instances and potentially leading to some ambiguous intra-cluster and inter-cluster distances~\citep{mishchuk2017working,zhang2022contrastive}, which is detrimental for SLU. Specifically, if clean manual transcripts are pushed away from their associated ASR transcripts while become closer to other sentences, the negative impact of ASR errors will be further exacerbated.

(3)~\textbf{They suffer from the problem of KL vanishing.} Inevitable label noise usually has a negative impact on the model~\cite{li2022past,cheng2023ssvmr}. Existing methods apply self-distillation to minimize Kullback–Leibler~(KL) divergence~\citep{kullback1951information} between the current prediction and the previous one to reduce the label noises in the training set. However, we find these methods suffer from the KL vanishing issue, which has been observed in other tasks~\citep{zhao2017learning}. KL vanishing can adversely affect the training of the model. Therefore, it is crucial to solve this problem to improve the performance.

In this paper, we propose \textbf{M}utual \textbf{L}earning and \textbf{L}arge-\textbf{M}argin \textbf{C}ontrastive \textbf{L}earning~(ML-LMCL), a novel framework to tackle above three issues. For the first issue, we propose a mutual learning paradigm. In fine-tuning, we train two SLU models on the manual and ASR transcripts, respectively. These two models are collaboratively trained and considered as peers, with the aim of iteratively learning and sharing the knowledge between the two models. Mutual learning allows effective dual knowledge transfer~\citep{liao2020multi,zhao2021mutual,zhu2021gaml}, which can improve the performance. For the second issue, our framework implements a large-margin contrastive learning to distinguish between intra-cluster and inter-cluster pairs. Specifically, we apply a distance polarization regularizer and penalize all pairwise distances within the margin region, which can encourage polarized distances for similarity determination and obtain a large margin in the distance space in an unsupervised way. For the third issue, following \citet{fu-etal-2019-cyclical}, we mitigate KL vanishing by adopting a cyclical annealing schedule. The training process is effectively split into many cycles. In each cycle, the coefficient of KL Divergence progressively increases from 0 to 1 during some iterations and then stays at 1 for the remaining iterations. Experiment results on three datasets SLURP, ATIS and TREC6~\citep{bastianelli2020slurp,hemphill1990atis,li2002learning,chang22c_interspeech} demonstrate that our ML-LMCL significantly outperforms previous best models and model analysis further verifies the advantages of our model.

The contributions of our work are four-fold:
\begin{itemize}[itemsep=2pt,topsep=0pt,parsep=0pt]
\item We propose ML-LMCL, which utilizes mutual learning to encourage the exchange of knowledge between the model trained on clean manual transcripts and the model trained on ASR transcripts. To the best of our knowledge, we make the first attempt to apply mutual learning to improve ASR robustness in SLU task.
\item To better distinguish between intra-cluster and inter-cluster pairs, we introduce a distance polarization regularizer to achieve large-margin contrastive learning. 
\item We adopt a cyclical annealing schedule to mitigate KL vanishing, which is neglected in the previous SLU approaches.
\item Experiments on three public datasets demonstrate that the proposed model achieves new state-of-the-art performance.
\end{itemize}
\section{Approach}
Our framework includes four elements:~(1)~Self-supervised contrastive learning with a distance polarization regularizer in pre-training.~(2)~Mutual learning between the model trained on clean manual transcripts and the model trained on ASR transcripts in fine-tuning.~(3)~Supervised contrastive learning with a distance polarization regularizer in fine-tuning.~(4)~Self-distillation with the cyclical annealing schedule in fine-tuning. 

\subsection{Self-supervised Contrastive Learning}\label{pre-train}
\begin{figure}[t]
\begin{minipage}[a]{1.0\linewidth}
  \centering
\centerline{\includegraphics[width=7.75cm]{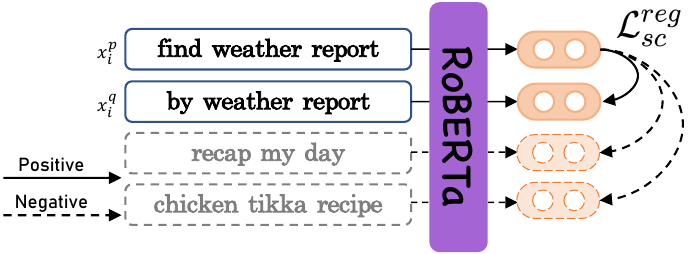}}
  \smallskip
\end{minipage}
\caption{The illustration of the pre-training stage. We apply large-margin self-supervised contrastive learning with paired transcripts. A positive pair consists of clean data and the associated ASR transcript.}
\label{fig:pre-train}
\end{figure}
\begin{figure*}[tb]
  \centering
\includegraphics[width=\linewidth]{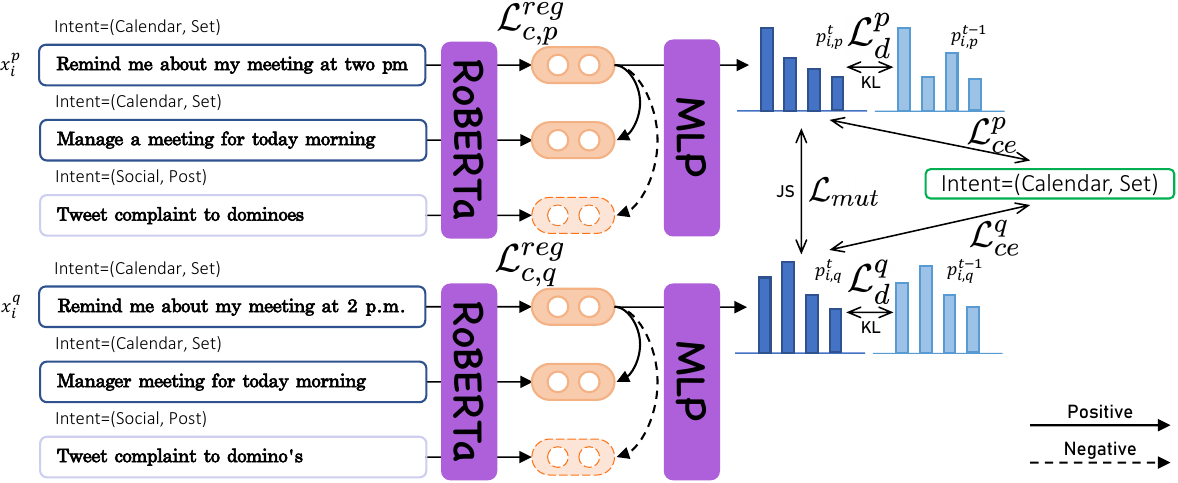}
  \vspace{0.2em}
  \caption{The illustration of the fine-tuning stage. Two networks on the clean manual transcripts and the ASR transcripts are  collaboratively trained via mutual learning~($\S\ref{mutual}$). Large-margin supervised contrastive learning~($\S\ref{super}$) and self-distillation~($\S\ref{self}$) are applied to further reduce the impact of ASR errors.}
  \label{fig:network}
\end{figure*}
Following \citet{chang22c_interspeech}, we utilize self-supervised contrastive learning in pre-training. Inspired by the success of pre-trained models~\cite{liu2022we,xin2022audio,chen2022hts,zhang2023speechgpt,cheng2023m,xin2023retrieval,zhang2023dub,xin2023detection,yang-etal-2023-multicapclip}, we continually train a RoBERTa~\citep{liu2019roberta} on spoken language corpus.

Given $N$ pairs of transcripts $\!\{(x^p_i, x^q_i)\}_{i=1 \dots N}$, where $x^p_i$ denotes a clean manual transcript and $x^q_i$ denotes its associated ASR transcript. As shown in Figure \ref{fig:pre-train}, we first utilize the pre-trained RoBERTa model and extract the representation from the last layer's \texttt{[CLS]} token $h_i^p$ for $x_i^p$ and $h_i^q$ for $x_i^q$:
\begin{align}
h_i^{p}&=\text{RoBERTa}(x_i^{p})\\
h_i^{q}&=\text{RoBERTa}(x_i^{q})
\end{align}

Then we utilize the self-supervised contrastive loss $\mathcal{L}_{sc}$~\citep{chen2020simple, gao2021simcse} to adjust the corresponding sentence representations:
\begin{equation}
\small
  \begin{split}
    \mathcal{L}_{sc} & = -\frac{1}{2N} \sum_{(h, h^{+}) \in P} \log \frac{e^{s(h, h^{+})/\tau_{sc}}} {\sum^B_{h' \neq h} e^{s(h, h') / \tau_{sc}}} \\
    & = -\mathbb{E}_P \Big[s(h, h^{+})/\tau_{sc}\Big] + \mathbb{E}\Big[\log\big(  {\sum^{B}_{h' \neq h} e^{s(h, h') / \tau_{sc}}} \big) \Big]
  \end{split}
  \label{eq3}
\end{equation}
where $P$ is composed of $2N$ positive pairs of either $(h^p_i, h^q_i)$ or $(h^q_i, h^p_i)$, $\tau_{sc}$ is the temperature hyper-parameter and \(s(\cdot,\cdot)\) denotes the cosine similarity function. However, conventional contrastive learning has a problem that semantically similar pairs are still pushed away~\citep{chen2021large}. It indiscriminately enlarges distances between all pairs of instances and may not be able to distinguish intra-cluster and inter-cluster correctly, which causes some similar instance pairs to still be pushed away. Moreover, it may discard some negative pairs and regard them as semantically similar pairs wrongly, even though their learning objective treat each pair of original instances as dissimilar. These problems result in the distance between the clean manual transcript and its associated ASR transcript not being significantly smaller than the distance between unpaired instance, which is detrimental to improving ASR robustness. Motivated by \citet{chen2021large}, we introduce a distance polarization regularizer to build a large-margin contrastive learning model. For simplicity, we further denote the following normalized cosine similarity:
\begin{equation}
    \mathcal{D}_{i j}=\left(1+s(h_{i}, h_{j})\right) / 2
\end{equation}
which measures the similarity between the pairs of  $(h_i, h_j)\in B$  with the real value  $\mathcal{D}_{i j} \in[0,1]$. We suppose that the matrix  $\mathbfcal{D}=\left\{\mathcal{D}_{i j} \in \mathbb{R}^{M \times M}\right\}$ where $M=2N$ denotes the total number of transcripts in $B$. $\mathbfcal{D}$ consists of distances $\mathcal{D}_{i j}$ and there exists $0<\delta^{+}<\delta^{-}<1$ where the intra-class distances are smaller than  $\delta^{+}$ while the inter-class distances are larger than  $\delta^{-}$. The proposed distance polarization regularizer $\mathcal{L}_{reg}$ is as follows:
\begin{equation}
\small
\mathcal{L}_{reg}=\left\|\min \left(\left(\mathbfcal{D}-\boldsymbol{\Delta}^{+}\right) \odot\left(\mathbfcal{D}-\boldsymbol{\Delta}^{-}\right), 0\right)\right\|_{1}
\end{equation}
where $\boldsymbol{\Delta}^{+}\!=\!\delta^{+} \times \mathbf{1}_{M \times M}$ and $\boldsymbol{\Delta}^{-}\!=\!\delta^{-} \times \mathbf{1}_{M \times M}$ are the threshold parameters and $\|\cdot\|_{1}$ denotes the $\ell_{1}$-norm. The region  $\left(\delta^{+}, \delta^{-}\right)$ $\subseteq[0,1]$  can be regarded as the large margin to discriminate the similarity of data pairs. $\mathcal{L}_{reg}$ can encourage the sparse distance distribution in the margin region $\left(\delta^{+}, \delta^{-}\right) $, because any distance $\mathcal{D}_{i j}$ fallen into the margin region  $\left(\delta^{+}, \delta^{-}\right)$ will increase $\mathcal{L}_{reg}$. Minimizing the regularizer $\mathcal{L}_{reg}$ will encourage more pairwise distances $\left\{\mathcal{D}_{i j}\right\}_{i, j=1}^{M}$ to distribute in the regions $\left[0, \delta^{+}\right]$ or  $\left[\delta^{-}, 1\right]$, and each data pair is adaptively separated into similar or dissimilar result. As a result, through introducing the regularizer, our framework can better distinguish between intra-cluster and inter-cluster pairs. 

Then the final large-margin self-supervised contrastive learning loss $\mathcal{L}_{sc}^{reg}$ is the weighted sum of self-supervised contrastive learning loss $\mathcal{L}_{sc}$ and the regularizer $\mathcal{L}_{reg}$, which is calculated as follows:
\begin{equation}
    \mathcal{L}_{sc}^{reg} = \mathcal{L}_{sc} + \lambda_{reg}\cdot \mathcal{L}_{reg}
\end{equation}
where $\lambda_{reg}$ is a hyper-parameter.
\subsection{Mutual Learning}
\label{mutual}
Previous work reveals that mutual learning can exploit the mutual guidance information between two models to improve their performance simultaneously~\citep{nie2018mutual,hong2021fine}. By mutual learning, we can obtain compact networks that perform better than those distilled from a strong but static teacher. In fine-tuning, we use the same pre-trained model in Sec.\ref{pre-train} to train two networks on the manual transcripts and the ASR transcripts, respectively. For a manual transcript $x^p_i$ and its associated ASR transcript $x^q_i$, the output probabilities $p_{i,p}^t$ and $p_{i,q}^t$ at the $t$-th epoch are as follows:
\begin{align}
p_{i,p}^t&=M_{\text{clean}}(x^p_i)\\
    p_{i,q}^t&=M_{\text{asr}}(x^q_i)
\end{align}
where $M_{\text{clean}}$ denotes the model trained on clean manual transcripts and $M_{\text{asr}}$ denotes the model trained on ASR transcripts.

We adopt Jensen-Shannon~(JS) divergence as the mimicry loss, with the aim of effectively encouraging the two models to mimic each other. The mutual learning loss $\mathcal{L}_{mut}$ in Figure \ref{fig:network} is as follows:
\begin{equation}
\mathcal{L}_{mut}=\sum_{i=1}^{N}JS(p_{i,p}^t\|p_{i,q}^t)
\end{equation}
\subsection{Supervised Contrastive Learning}
\label{super}
We also apply supervised contrastive learning in fine-tuning by using label information. The pairs with the same label are regarded as positive samples and the pairs with different labels are regarded as negative samples. The embeddings of positive samples are pulled closer while the embeddings of negative samples are pushed away~\citep{jian-etal-2022-contrastive,zhou2022knn}. We utilize the supervised contrastive loss $\mathcal{L}_{c}^{p}$ for the model trained on manual transcripts and $\mathcal{L}_{c}^{q}$ for the model trained on ASR transcripts to encourage the learned representations to be aligned with their labels:

\begin{small}
\begin{align}
        \mathcal{L}_{c}^{p}\!=\!-\frac{1}{N} \cdot \sum^N_{i=1} \sum^N_{j\neq i} 1_{y_i^p=y_j^p} \log \frac{ e^{s(h_i^p, h_j^p)/\tau_{c}}}{\sum^N_{k\neq i} e^{s(h_i^p, h_k^p) / \tau_{c}}}\\
        \mathcal{L}_{c}^{q}\!=\!-\frac{1}{N} \cdot \sum^N_{i=1} \sum^N_{j\neq i} 1_{y_i^q=y_j^q} \log \frac{ e^{s(h_i^q, h_j^q)/\tau_{c}}}{\sum^N_{k\neq i} e^{s(h_i^q, h_k^q) / \tau_{c}}}    
\end{align}
\end{small}

\noindent where $y_i^p\!=\!y_j^p$ denotes the labels of $h_i^p$ and $h_j^p$ are the same, $y_i^q\!=\!y_j^q$ denotes the label of $h_i^q$ and $h_j^q$ are the same and $\tau_{c}$ is the temperature hyper-parameter.

Like Sec.\ref{pre-train}, we also use distance polarization regularizers $\mathcal{L}_{reg}^p$ and $\mathcal{L}_{reg}^q$ to enhance the generalization ability of contrastive learning algorithm:

\begin{small}
\begin{align}
\mathcal{L}_{reg}^p=\left\|\min \left(\left(\mathbfcal{D}^p-\boldsymbol{\Delta}^{+}\right) \odot\left(\mathbfcal{D}^p-\boldsymbol{\Delta}^{-}\right), 0\right)\right\|_{1}\\
\mathcal{L}_{reg}^q=\left\|\min \left(\left(\mathbfcal{D}^q-\boldsymbol{\Delta}^{+}\right) \odot\left(\mathbfcal{D}^q-\boldsymbol{\Delta}^{-}\right), 0\right)\right\|_{1}  
\end{align}
\end{small}

\noindent where $\mathbfcal{D}^p$ denotes the matrix consisting of pairwise distances on the clean manual transcripts and $\mathbfcal{D}^q$ denotes the matrix on the ASR transcripts.

The large-margin supervised contrastive learning loss $\mathcal{L}_{c,p}^{reg}$ and $\mathcal{L}_{c,q}^{reg}$ in Figure \ref{fig:network} are as follows:
\begin{align}
\mathcal{L}_{c,p}^{reg}&=\mathcal{L}_c^p+\lambda_{reg}^p \mathcal{L}^p_{reg}\\
\mathcal{L}_{c,q}^{reg}&=\mathcal{L}_c^q+\lambda_{reg}^q\mathcal{L}^q_{reg}    
\end{align}
where $\lambda_{reg}^p$ and $\lambda_{reg}^q$ are two hyper-parameters.

The final large-margin supervised contrastive learning loss $\mathcal{L}_{c}^{reg}$ is as follows:
\begin{equation}
\mathcal{L}_{c}^{reg}\!=\mathcal{L}_{c,p}^{reg}+\mathcal{L}_{c,q}^{reg}
\end{equation}
\subsection{Self-distillation}
\label{self}
To further reduce the impact of ASR errors, we apply a self-distillation method. We try to regularize the model by minimizing Kullback–Leibler~(KL) divergence~\citep{kullback1951information,he2022weighted} between the current prediction and the previous one~\citep{liu2020fastbert,liu2021noisy}. For the manual transcript $x_i^p$ and its corresponding label $y_i^p$, $p^{t}_{i,p}=P(y_i^p|x_i^p, t)$ denotes the probability distribution of $x_i^p$ at the $t$-th epoch, and $p^{t}_{i,q}=P(y_i^q|x_i^q, t)$ denotes the probability distribution of $x_i^q$ at the $t$-th epoch. The loss functions $\mathcal{L}_d^{p}$ and $\mathcal{L}_d^{q}$ of self-distillation in Figure \ref{fig:network} are formulated as:
\begin{align}
    \mathcal{L}_d^{p}\!&=\!\frac{1}{N}\sum^N_{i=1} \tau^2_{d}KL\Big(\frac{p^{t-1}_{i,p}}{\tau_{d}} \| \frac{p^{t}_{i,p}}{\tau_{d}}\Big)\\
    \mathcal{L}_d^{q}\!&=\!\frac{1}{N}\sum^N_{i=1} \tau^2_{d}KL\Big(\frac{p^{t-1}_{i,q}}{\tau_{d}} \| \frac{p^{t}_{i,q}}{\tau_{d}}\Big)
\end{align}
where $\tau_{d}$ is the temperature to scale the smoothness of two distributions, note that $p^0_{i,p}$ is the one-hot vector of label $y_i^p$ and $p^0_{i,q}$ is that of label $y_i^q$.

Then the final self-distillation loss $\mathcal{L}_{d}$ is the sum of two loss functions $\mathcal{L}_d^{p}$ and $\mathcal{L}_d^{q}$:
\begin{equation}
    \mathcal{L}_{d}=\mathcal{L}_d^{p}+\mathcal{L}_d^{q}
\end{equation}

\subsection{Training Objective}
\paragraph{Pre-training}
Following \cite{chang22c_interspeech}, the pre-training loss $\mathcal{L}_{pt}$ is the weighted sum of the large-margin self-supervised contrastive learning loss $\mathcal{L}_{sc}^{reg}$ and an MLM loss $\mathcal{L}_{mlm}$:
\begin{equation}
  \mathcal{L}_{pt} = \lambda_{pt}\mathcal{L}_{sc}^{reg}+(1-\lambda_{pt})\cdot \mathcal{L}_{mlm}
\end{equation}
where $\lambda_{pt}$ is the coefficient balancing the two tasks.
\paragraph{Fine-tuning}
Following \citet{haihong2019novel,HAN,cheng23b_interspeech,zhu2023icassp}, the intent detection objective is:
\begin{align}
\mathcal{L}_{ce}^p&=-\sum_{i=1}^{N} y_{i}^{p} \log {p}_{i,p}^{t}\\
\mathcal{L}_{ce}^q&=-\sum_{i=1}^{N} y_{i}^{q} \log {p}_{i,q}^{t}\\
\mathcal{L}_{ce}&=\mathcal{L}_{ce}^p+\mathcal{L}_{ce}^q
\end{align}

The final fine-tuning loss $\mathcal{L}_{ft}$ is the weighted sum of cross-entropy loss $L_{ce}$, mutual learning loss $\mathcal{L}_{mut}$, large-margin supervised contrastive learning loss $\mathcal{L}_c^{reg}$ and self-distillation loss $\mathcal{L}_d$:
\begin{equation}
\mathcal{L}_{ft}=\mathcal{L}_{ce}+\alpha \mathcal{L}_{mut} + \beta \mathcal{L}_c^{reg} + \gamma \mathcal{L}_d
\label{final}
\end{equation}
where $\alpha$, $\beta$, $\gamma$ are the trade-off hyper-parameters.

However, directly using KL divergence for self-ditillation loss may suffer from the vanishing issue. To mitigate KL vanishing issue, we adopt a cyclical annealing schedule, which is also applied for this purpose in \citet{fu-etal-2019-cyclical,zhao2021mutual}. Concretely, $\gamma$ in Eq.\ref{final} changes periodically during training iterations, which is described by Eq.\ref{annealing}:
\begin{align}
\label{annealing}
    \gamma&=\left\{\begin{array}{lr}
\frac{r}{R C}, & r\leqslant RG \\
1, & r>RG
\end{array}\right.\\
r&=mod(t-1,G)
\end{align}
where $t$ represents the current training iteration and $R$ and $G$ are two hyper-parameters.
\section{Experiments}
\subsection{Datasets and Metrics}
Following \citet{chang22c_interspeech}, we conduct the experiments on three benchmark datasets\footnote{SLURP is available at \url{https://github.com/MiuLab/SpokenCSE}, and ATIS and TREC6 are available at \url{https://github.com/Observeai-Research/Phoneme-BERT}.}:~SLURP, ATIS and TREC6~\citep{hemphill1990atis,li2002learning,chang22c_interspeech,bastianelli2020slurp}, whose statistics are shown in Table \ref{tab:datasets}.
\begin{table}[ht]
  \centering
\setlength{\tabcolsep}{1.1mm}{
  \begin{tabular}{ l r r r r}
    \toprule
    \multicolumn{1}{c}{\textbf{Dataset}} & 
        \multicolumn{1}{c}{\textbf{\#Class}} &
        \multicolumn{1}{c}{\textbf{Avg. Length}} &
        \multicolumn{1}{c}{\textbf{Train}} & 
        \multicolumn{1}{c}{\textbf{Test}} \\
    \midrule
    SLURP & \(18\times46\) & 6.93 & 50,628 & 10,992\\
    ATIS & 22 & 11.14 & 4,978 & 893  \\
    TREC6 & 6 & 8.89 & 5,452 & 500 \\
   \bottomrule
  \end{tabular}}
\caption{The statistics of all datasets. The \textit{test} set of SLURP is sub-sampled.}
\label{tab:datasets}
\end{table}

SLURP is a challenging SLU dataset with various domains, speakers, and recording settings. An intent of SLURP is a (scenario, action) pair, the joint accuracy is used as the evaluation metric and the prediction is regarded correct only when both the scenario and action are correctly predicted. The ASR transcripts are obtained by Google Web API.

\begin{table*}[ht]
\centering
\resizebox{\textwidth}{!}{
\fontsize{8}{10}\selectfont
\label{tab:main_results}
\begin{tabular}{lcccccc}
\toprule
\multirow{2.5}{*}{Model}&
\multicolumn{3}{c}{w/o manual transcripts}&\multicolumn{3}{c}{w/ manual transcripts}\cr
\cmidrule(lr){2-4} \cmidrule(lr){5-7}
&SLURP&ATIS&TREC6&SLURP&ATIS&TREC6\cr
\midrule
RoBERTa~\citep{liu2019roberta}& 83.97& 94.53& 84.08& 84.42& 94.86& 84.54\cr
Phoneme-BERT~\citep{sundararaman2021phoneme}& 83.78& 94.83& 85.96& 84.16&95.14& 86.48\cr
SimCSE~\citep{gao2021simcse}& 84.47& 94.07& 84.92& 84.88& 94.32& 85.46\cr
SpokenCSE~\citep{chang22c_interspeech}& 85.26& 95.10& 86.36& 85.64& 95.58& 86.82\cr
\midrule
ML-LMCL&\textbf{88.52}$^{\dag}$&\textbf{96.52}$^{\dag}$&\textbf {89.24}$^{\dag}$& \textbf{89.16}$^{\dag}$&\textbf{97.21}$^{\dag}$&\textbf{89.96}$^{\dag}$\cr
\bottomrule
\end{tabular}}
\caption{Accuracy results on three datasets. $\dag$ denotes ML-LMCL obtains statistically significant improvements over baselines with p < 0.01. "w/o manual transcripts" denotes clean manual transcripts are not used in fine-tuning, \textit{i.e.} the loss functions associated with clean manual transcripts are set to 0, including $\mathcal{L}_{ce}^p$, $\mathcal{L}_{mut}$, $\mathcal{L}_{c,p}^{reg}$, and $\mathcal{L}_{d}^p$. "w/ manual transcripts" denotes clean manual transcripts are used in fine-tuning.}
\label{results}
\end{table*}
ATIS and TREC6 are two SLU datasets for flight reservation and question classification respectively. We use the synthesized text released by Phoneme-BERT~\citep{sundararaman2021phoneme}. 
We adopt accuracy as the evaluation metric for intent detection.

\subsection{Implementation Details}
We perform pre-training on the model for 10,000 steps using the batch size of 128 for each dataset. Afterward, we fine-tune the entire model for up to 10 epochs, utilizing a batch size of 256 to mitigate overfitting.
The mask ratio of MLM is set to 0.15, $\tau_{sc}$ is set to 0.2, $\delta^{+}$ is set to 0.2, $\delta^{-}$ is set to 0.5, $\lambda_{reg}$ is set to 0.1, $\tau_{c}$ is set to 0.2, $\lambda_{reg}^p$ is set to 0.15, $\lambda_{reg}^q$ is set to 0.15, $\tau_{d}$ is set to 5, $\lambda_{pt}$ is set to 0.5, $\alpha$ is set to 1, $\beta$ is set to 0.1, $R$ is set to 0.5, and $G$ is set to 5000. The reported scores are averaged over 5 runs. During both the pre-training and fine-tuning stages, we employ the Adam optimizer~\citep{kingma2014adam} with $\beta_1=0.9$ and $\beta_2=0.98$. Additionally, we incorporate 4,000 warm-up updates to optimize the model parameters. The training process typically spans a few hours and is conducted on an Nvidia Tesla-A100 GPU.
\subsection{Main Results}
We compare our ML-LMCL with four baselines, including RoBERTa~\citep{liu2019roberta}, Phoneme-BERT~\citep{sundararaman2021phoneme}, SimCSE~\citep{gao2021simcse}, and SpokenCSE~\citep{chang22c_interspeech}. The performance comparison of ML-LMCL and the baselines are shown in Table \ref{results}, from which we have the following observations:

(1)~Our ML-LMCL approach consistently yields improvements across all tasks and datasets. This can be attributed to the mutual guidance achieved between the models trained on manual and ASR transcripts, enabling them to share knowledge effectively. In addition, the adoption of large-margin contrastive learning encourages the model to distinguish between intra-cluster and inter-cluster pairs more accurately, minimizing the separation of semantically similar pairs. To overcome the issue of KL vanish, we apply a cyclical annealing schedule, which enhances the model's robustness. Notably, even when manual transcripts are not utilized, our approach outperforms SpokenCSE, further highlighting the efficacy of the large-margin contrastive learning and the cyclical annealing schedule in enhancing ASR robustness in SLU.

(2)~In contrast, the more significant improvement observed on the SLURP dataset could be attributed to its inherent difficulty compared to the ATIS and TREC6 datasets. SLURP presents a challenge in SLU as its intents consist of (scenario, action) pairs, and a prediction is deemed correct only if both the scenario and action are accurately predicted. Previous approaches using the conventional contrastive learning methods have struggled to achieve precise alignment between ASR transcripts and their corresponding manual transcripts. Consequently, due to ASR errors, it is common for one of the two components of an intent to be incorrectly predicted. Our ML-LMCL approach addresses these limitations of conventional contrastive learning, resulting in improved alignment and performance.
\subsection{Analysis}
To verify the advantages of ML-LMCL from different perspectives, we use clean manual transcripts and conduct a set of ablation experiments. The experimental results are shown in Table \ref{tab:ablation}.
\begin{table}[ht] 
	\centering
	\begin{adjustbox}{width=0.48\textwidth}
		\begin{tabular}{l|ccc}
			\hline
			\multirow{2}{*}{\textbf{Model}} & \multicolumn{3}{c}{w/ manual transcripts}  \\
			\cline{2-4}
			& SLURP & ATIS & TREC6 
			\\
			\hline
ML-LMCL   &  \textbf{89.16} &\textbf{97.21} &\textbf{89.96}  \\\hline
w/o $\mathcal{L}_{mut}$  &  88.68 ($\downarrow$0.48)&  96.83  ($\downarrow$0.38)    &   89.52  ($\downarrow$0.44)    \\
w/o $\mathcal{L}_{reg}$  &  88.92 ($\downarrow$0.24)&  96.98  ($\downarrow$0.23)    &   89.77  ($\downarrow$0.19)    \\
w/o $\mathcal{L}_{reg}^p$ \& $\mathcal{L}_{reg}^q$ &  88.75 ($\downarrow$0.41)&  96.92  ($\downarrow$0.29)    &   89.74  ($\downarrow$0.22)    \\
w/o cyc &  88.98 ($\downarrow$0.18) &  97.08  ($\downarrow$0.13)  & 89.85 ($\downarrow$0.11)   \\\hdashline
w/o $L_{mut}$ + bsz$\uparrow$ &  88.72 ($\downarrow$0.44) &  96.92  ($\downarrow$0.29)  & 89.65 ($\downarrow$0.31)  \\\hdashline
w/ $\mathcal{L}_{soft}$ &  89.12 ($\downarrow$0.04) &  97.18  ($\downarrow$0.03)  & 89.92 ($\downarrow$0.04)  \\
\hline
		\end{tabular}	
 \end{adjustbox}
	\caption{Results of the ablation experiments when using clean manual transcripts.
	} 
\label{tab:ablation}
\end{table}
\subsubsection{Effectiveness of Mutual Learning}
One of the core contributions of ML-LMCL is mutual learning, which allows the two models trained on manual and ASR transcripts learn from each other. To verify the effectiveness of mutual learning, we remove mutual learning loss and refer it to \textit{w/o $\mathcal{L}_{mut}$} in Table \ref{tab:ablation}. We observe that accuracy drops by 0.48, 0.38 and 0.44 on SLURP, ATIS and TREC6, respectively. Contrastive learning benefits more from larger batch size because larger batch size provides more negative examples to facilitate convergence~\citep{chen2020simple}, and many attempts have been made to improve the performance of contrastive learning by increasing batch size indirectly~\citep{he2020momentum,chen2020improved}. Therefore, to verify that the proposed mutual learning rather than the indirectly boosted batch sizes works, we double the batch size after removing mutual learning loss and refer it to \textit{w/o $L_{mut}$ + bsz$\uparrow$}. The results demonstrate that despite the boosted batch size, it still performs worse than ML-LMCL, which indicates that the observed enhancement primarily arises from mutual learning approach, rather than from the increased batch size.
\subsubsection{Effectiveness of Distance Polarization Regularizer}
To verify the effectiveness of distance polarization regularizer, we also remove distance polarization regularizer in pre-training and fine-tuning, which is named as \textit{w/o $\mathcal{L}_{reg}$} and \textit{w/o $\mathcal{L}_{reg}^p$} \& \textit{$\mathcal{L}_{reg}^q$}, respectively. When $\mathcal{L}_{reg}$ is removed, the accuracy drops by 0.24, 0.23 and 0.19 on SLURP, ATIS and TREC6, respectively. And when $\mathcal{L}_{reg}^p$ and $\mathcal{L}_{reg}^q$ are removed, the accuracy drops by 0.41, 0.29 and 0.22 on SLURP, ATIS and TREC6. The results demonstrate that distance polarization regularizer can alleviate the negative impact of conventional contrastive learning. Furthermore, the drop in accuracy is greater when fine-tuning than when pre-training. We believe that the reason is that supervised contrast learning in fine-tuning is easier to be affected by label noise than unsupervised contrast learning in pre-training. As a result, more semantically similar pairs are incorrectly pushed away in fine-tuning when the regularizer is removed. 

\citet{chang22c_interspeech} also proposes a self-distilled soft contrastive learning loss to relieve the negative effect of noisy labels in supervised contrastive learning. However, we believe that the regularizer can also effectively reduce the impact of label noise. Therefore, our ML-LMCL does not include another module to tackle the problem of label noise. To verify this, we augument ML-LMCL with the self-distilled soft contrastive learning loss, which is termed as \textit{w/ $\mathcal{L}_{soft}$}. We can observe that not only $\mathcal{L}_{soft}$ does not bring any improvement, it even causes performance drops, which proves that the distance polarization regularizer can indeed reduce the impact of label noise.
\subsubsection{Effectiveness of Cyclical Annealing Schedule}
\begin{figure}[t]
\begin{minipage}[a]{1.0\linewidth}
  \centering
\centerline{\includegraphics[width=7.75cm]{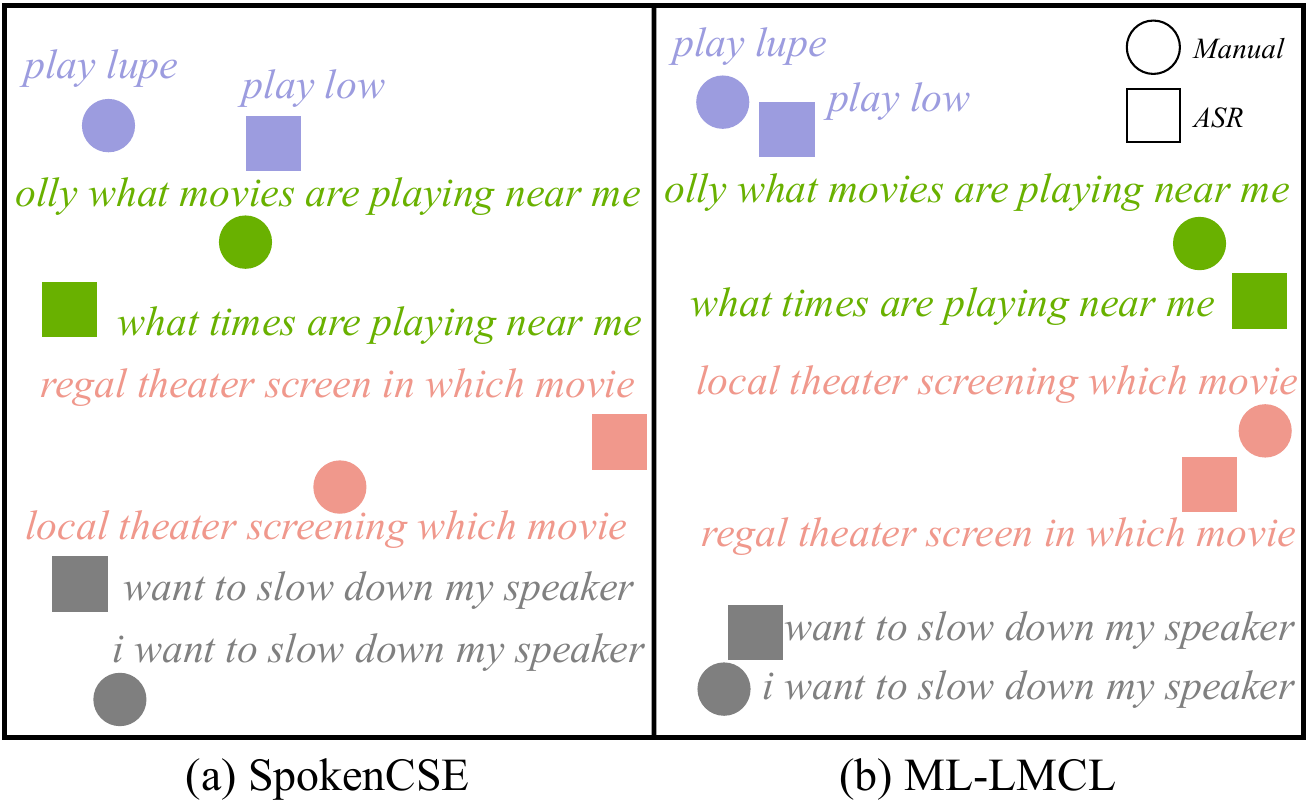}}
  \smallskip
\end{minipage}
\caption{Visualization of representations of manual transcripts and ASR transcripts. We visualize the representations by reducing the dimension with Principal Component Analysis~(PCA)~\citep{abdi2010principal}. The circle and square in the same color means the corresponding manual and ASR transcriptions are associated.}
\label{fig:visual}
\end{figure}
We also remove cyclical annealing schedule and relate it to \textit{w/o cyc}. We observe that the accuracy drops by 0.18, 0.13 and 0.11 on SLURP, ATIS and TREC6, respectively, which demonstrates that the cyclical annealing schedule also plays an important role in enhancing the performance by mitigating the problem of KL vanishing. 

\subsection{Visualization}
To gain a deeper understanding of the impact and contribution of mutual learning and large-margin contrastive learning, we present a visualization of an example from the SLURP dataset in Figure \ref{fig:visual}. In this example, we compare the manual transcripts \textit{local theater screening which movie''} and \textit{olly what movies are playing near me''}, which share the same intent. In our ML-LMCL approach, the representations of these transcripts along with their associated ASR transcripts remain closely clustered. Conversely, in SpokenCSE, there is a greater separation between their representations, further illustrating that our method effectively aligns ASR and manual transcripts with high accuracy and minimizes the pushing apart of similar pairs.
\section{Related work}
\paragraph{Error-robust Spoken Language Understanding} SLU usually suffers from ASR error propagation and this paper focus on improving ASR robustness in SLU. \citet{chang22c_interspeech} makes the first attempt to use contrastive learning to improve ASR robustness with only textual information. Following \citet{chang22c_interspeech}, this paper only focuses on intent detection in SLU. Intent detection is usually formulated as an utterance classification problem. 
As a large number of pre-trained models achieve surprising results across various tasks~\cite{dong2022survey,yang2023implicit,cheng2023towards,zhu2023,yang2023zeronlg}, some BERT-based~\citep{devlin2019bert} pre-trained work has been explored in SLU where the representation of the special token \texttt{[CLS]} is used for intent detection. In our work, we adopt RoBERTa and try to learn the invariant representations between clean manual transcripts and erroneous ASR transcripts.

\paragraph{Mutual Learning} Our method is motivated by the recent success in mutual learning.
Mutual learning is an effective method which trains two models of the same architecture simultaneously but with different initialization and encourages them to learn collaboratively from each other.
Unlike knowledge distillation~\citep{hinton2015distilling}, mutual learning doesn't need a powerful teacher network which is not always available.
Mutual learning is first proposed to leverage information from multiple models and allow effective dual knowledge transfer in image processing tasks~\citep{zhang2018deep,zhao2021mutual}. Based on this, \citet{wu2019mutual-semi} utilizes mutual learning to capture complementary features in semi-supervised classification.
In NLP area, \citet{zhao2021mutual} utilizes mutual learning for speech translation to transfer knowledge between a speech translation model and a machine translation model. In our work, we apply a mutual learning framework to transfer knowledge between the models trained on manual and ASR transcripts. 

\paragraph{Contrastive learning}
Contrastive learning aims at learning example representations by minimizing the distance between the positive pairs in the vector space and maximizing the distance between the negative pairs~\citep{saunshi2019theoretical,liang2022jointcl,liu2022twin}, which is first proposed in the field of computer vision~\citep{chopra2005learning,schroff2015facenet,sohn2016improved,chen2020simple,wang2021understanding}. In the NLP area, contrastive learning is applied to learn sentence embeddings~\citep{giorgi2021declutr,yan2021consert}, translation~\citep{pan2021contrastive,ye-etal-2022-cross} and summarization~\citep{wang2021contrastive, cao2021cliff}. 
Contrastive learning is also used to learning a unified representation of image and text~\citep{dong2019unified,zhou2020unified,li2021unimo}. 
Recently, \citet{chen2021large} points that conventional contrastive learning algorithms are still not good enough since they fail to maintain a large margin in the distance space for reliable instance discrimination so that semantically similar pairs are still pushed away. 
Inspired by this, we add a similar distance polarization regularizer as \citet{chen2021large} to address this issue. To the best of our knowledge, we are the first to introduce the idea of large-margin contrastive learning to the SLU task.
\section{Conclusion}
In this work, we propose a novel framework ML-LMCL for improving ASR robustness in SLU. We utilize mutual learning and introduce the distance polarization regularizer. Moreover, cyclical annealing schedule is utilized to mitigate KL vanishing. Experimental results and analysis on three benchmark datasets show that it significantly outperforms previous SLU models whether the clean manual transcriptions are available in fine-tuning or not. Future work will focus on improving ASR robustness with only clean manual transcriptions.
\section*{Limitations}
By applying mutual learning, introducing distance polarization regularizer and utilizing cyclical annealing schedule, ML-LMCL achieves significant improvement on three benchmark datasets. Nevertheless, we summarize two limitations for further discussion and investigation of other researchers:

(1)~ML-LMCL still requires the ASR transcripts in fine-tuning to align with the target inference scenario. However, the ASR transcripts may not always be readily available due to the constraint of ASR systems and privacy concerns. In the future work, we will attempt to further improve ASR robustness without using any ASR transcripts.

(2)~The training and inference runtime of ML-LMCL is larger than that of baselines. We attribute the extra cost to the fact that ML-LMCL has more parameters than baselines. In the future work, we plan to design a new paradigm with fewer parameters to reduce the requirement for GPU resources.

\section*{Acknowledgements}
We thank all anonymous reviewers for their constructive comments. This paper was partially supported by Shenzhen Science \& Technology Research Program (No: GXWD20201231165807007-20200814115301001) and NSFC (No: 62176008).

% \section*{Ethics Statement}
% Scientific work published at ACL 2023 must comply with the ACL Ethics Policy.\footnote{\url{https://www.aclweb.org/portal/content/acl-code-ethics}} We encourage all authors to include an explicit ethics statement on the broader impact of the work, or other ethical considerations after the conclusion but before the references. The ethics statement will not count toward the page limit (8 pages for long, 4 pages for short papers).

% Entries for the entire Anthology, followed by custom entries
\bibliography{anthology,custom}
\bibliographystyle{acl_natbib}

\appendix

\end{document}